\documentclass{article}

\PassOptionsToPackage{numbers, compress}{natbib}

\usepackage[preprint]{neurips_2021}

\usepackage[utf8]{inputenc} 
\usepackage[T1]{fontenc}    
\usepackage{hyperref}       
\usepackage{url}            
\usepackage{booktabs}       
\usepackage{amsfonts}       
\usepackage{amsmath}
\usepackage{nicefrac}       
\usepackage{microtype}      
\usepackage{xcolor}         
\usepackage{graphicx}
\usepackage{multirow}
\usepackage[ruled,vlined]{algorithm2e}
\usepackage{float}

\SetCommentSty{mycommfont}

\title{VSGM - Enhance robot task understanding ability through visual semantic graph}

\author{%
  Cheng Yu Tsai\\
  Department of Computer Science\\
  National Central University\\
  \texttt{roy1997860328@gmail.com} \\
  \And
  Mu-Chun Su \\
  Department of Computer Science\\
  National Central University\\
  \texttt{muchun.su@g.ncu.edu.tw} \\
}

\begin{document}

\maketitle

\begin{abstract}
  In recent years, developing AI for robotics has raised much attention. The interaction of vision and language of robots is particularly difficult. We consider that giving robots an understanding of visual semantics and language semantics will improve inference ability. In this paper, we propose a novel method-VSGM (Visual Semantic Graph Memory), which uses the semantic graph to obtain better visual image features, improve the robot's visual understanding ability. By providing prior knowledge of the robot and detecting the objects in the image, it predicts the correlation between the attributes of the object and the objects and converts them into a graph-based representation; and mapping the object in the image to be a top-down egocentric map. Finally, the important object features of the current task are extracted by Graph Neural Networks. The method proposed in this paper is verified in the ALFRED (Action Learning From Realistic Environments and Directives) dataset. In this dataset, the robot needs to perform daily indoor household tasks following the required language instructions. After the model is added to the VSGM, the task success rate can be improved by 6\textasciitilde10 \%.\footnote{Source code: https://github.com/roy402/VSGM}

\end{abstract}

\section{Introduction}

In recent years, there have been many applications using deep learning in the research of robot arms and robot interactions, which are usually an integrated method. From the robot's vision and language command reception and understanding to the final decision-making interaction, it has deepened the difficulty of its application. The dataset provided by robot instructions tends to become more and more abundant. 

Vision Language Navigation are topics that have attracted more and more attention in the subfield of robotics. The robot receives language commands and performs tasks in the limited space of action in the environment, which involves the interaction of semantic understanding and computer vision. In addition to \citep{misra2018mapping} providing a combination of visual and language instructions to operate tasks in a simple simulation environment; \citep{savva2019habitat} providing API to perform destination navigation tasks in an indoor simulation environment \citep{chang2017matterport3d}, resulting in many Deep Learning and Reinforcement Learning research \citep{wijmans2019dd,gordon2019splitnet}; Room-to-Room (R2R)\citep{anderson2018vision,thomason2020vision} provides language navigation instructions, instructing the robot to follow the language commands and navigate in the indoor simulation environment \citep{majumdar2020improving,fried2018speaker,wang2020environment,wang2019reinforced,tan2019learning}. \citep{krantz2020beyond} optimized the environment of \citep{anderson2018vision}, which eliminated the limitation of robot movement with the sparse graph of panoramas, and achieved a more realistic continuous simulation environment for experiments.

There are also many papers using AI2-THOR \citep{kolve2017ai2} 3D life simulation environment to conduct research \citep{jain2020cordial,lv2020improving,yang2018visual,nguyen2019reinforcement,qiulearning,du2020learning,kazemi2020optimistic,wortsman2019learning,gan2020look}. ALFRED \citep{shridhar2020alfred}(Action Learning From Realistic Environments and Directives) also proposed complex instructions for daily tasks in AI2-THOR, trains robots, and tests the feasibility of algorithms. It provides action operation datasets and language commands displayed by experts, including 13 types of basic action operations, main task instructions, and step-by-step instructions that are used to train the robot to complete tasks under human instructions. The baseline model provided in ALFRED only achieved a 4 \% task success rate in the test experiment, which has room for improvement.

In visual language tasks, we believe that the key factors that affect the success rate of the robot (agent) are: visual analysis ability, understanding of the purpose of the task, following instructions, memorizing, and comprehensive information for planning and execution. In addition to the type of object information, the visual image also includes implicit object information, object status, object position, the relationship between objects, and the current environment, etc. This information can be analyzed and judged based on human prior knowledge. In this paper, we propose two methods of the semantic graph and object mapping. We obtain and record object information to improve the agent's environmental understanding ability, thereby increasing the task success rate.

The following are our contributions:
\begin{itemize}
  \item 1. We propose a visual feature method that expresses the relationship between objects in the field of vision by the semantic graph. By extracting the object associations, object attributes, and storage objects in the image, they can be provided to the agent to achieve better visual semantic understanding. In addition, the semantic graph continuously updated achieves the continuous evolution of the task.
  \item 2. We propose a Spatial Semantic Map stored as a graph. Spatial Semantic Map is a bird's-eye map, which maps objects to the map. Graph Neural Networks embeds the map to obtain the features on the spatial information, in which the nodes are represented as objects. As a result, it can increase the agent's navigation and memory of the spatial information.
  \item 3. In the ALFRED dataset, the experimental results prove that the semantic graph and Spatial Semantic Map can help improve the success rate of the task; and increase the interpretability of the robot operation, which makes it easier to optimize the agent model.
\end{itemize}

\section{Related Work}
\label{gen_inst}

There are many related studies and practices in the ALFRED dataset. \citep{shridhar2020alfworld} allows the agent to change from a complicated operation instruction to a simple operation instruction. Using the additional text training environment lets the agent have a better high-level language cognition; MOCA \citep{singh2020moca} disassembled the task issue into multiple modules of vision, language, object location, ambiguous text, and plan execution; \citep{saha2021modular} disassembles task issues into multiple modules for vision, language, object location, unrecognizable text, and plan execution; \citep{storks2021we} proposes to expand the field of view, object position estimation, and strengthen the robot's navigation capabilities; \citep{zhu2020hierarchical,xu2019regression,zhu2017visual} uses planning capabilities similar to humans to simplify the difficulty of training tasks. However, only \citep{shridhar2020alfworld} performs additional abstract understanding training on task instructions. Moreover, these papers did not attempt to extract semantic features specifically for the images of the task. Therefore, our research focuses on the results of visual semantic feature extraction.

\begin{figure}[t]
  \centering
  \includegraphics[width=0.7\linewidth]{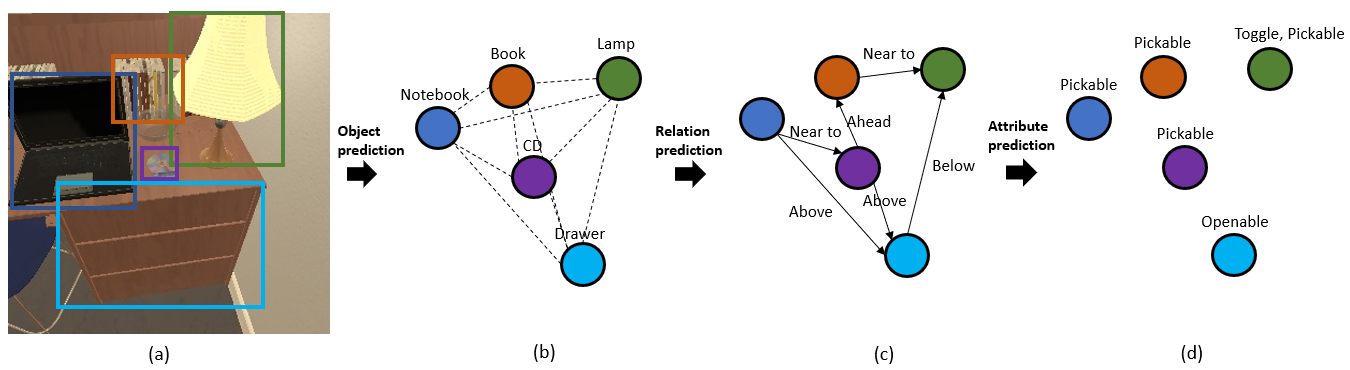}
  \caption{Scene Graph Generation first obtains the objects in the image by using MaskRCNN. Then express the object as a node, predict the relevance of the edge between the nodes, if there is relevance, use the edge to connect, and mark what kind of relevance this is.}
  \label{sgg}
\end{figure}

Robots can perform more effective inferences by applying spatial information. In recent years, there have been many methods including \textbf{topology graphs, mapping, and scene graph}. \citep{chaplot2020neural,chen2020topological,beeching2020learning} use nodes to represent the current panorama and then connects each node in series to represent the environment. Although the panorama image contains dozens of objects, it can't identify the object features, object information, and location information. \citep{chaplot2020learning,chen2020learning} combines SLAM (Simultaneous localization and mapping) to perform the task of navigating to the destination. \citep{cartillier2020semantic,shen2021spatial,wani2020multion,seymour2021maast} uses single-layer or multi-layer spatial mapping to map the detected objects and depth images to a top-down map. The approach increases the robot's spatial understanding capabilities. All the mapping and topological methods mentioned above focus on navigation, navigation to specific objects, and exploration. However, the execution of intricate tasks includes semantic understanding, navigation, exploration, object detection, object function understanding, and object manipulation. We believe that the navigation space information generated by SLAM and the image features of the topology has lack object information, and the meticulous operation will be limited.
To solve the daily life tasks, the mapping method proposed by \citep{saha2021modular}, applied to the ALFRED dataset, is a good research direction. However, the mapping method only includes Unknown, Navigable Space, Target, Obstacle, and object type. There are many types of objects, and only relying on the one-hot encode feature of the object types makes it difficult for the robot to analyze the relationship between the objects and have a good inference ability.

\textbf{Scene Graph} stores and describes the location of objects in space \citep{rosinol20203d,armeni20193d}. \citep{liao2020tsm,kim20193} saves objects as 3D scene graphs, which can record detailed spatial information. The robot can use more common models in the visual processing of images, such as Resnet\citep{he2016deep}, YOLO\citep{redmon2016you}, MaskRCNN\citep{he2017mask}, etc. According to the characteristics of CNN, it will focus on local features and discard spatial information. In contrast, \textbf{Scene Graph Generation} (SGG) (Figure~\ref{sgg}) \citep{li2018factorizable,yang2018graph,tang2020unbiased} can provide more spatial information for robots to make judgments. It not only combines ordinary visual processing methods but also gives relationships between objects.

\textbf{Semantic Graph} is different from Scene Graph. It uses detected objects as nodes, which node features are image features, and uses Graph Neural  Networks to extract features from nodes. In the task of navigating to the target, \citep{qiulearning,yang2018visual} use the current image to generate a Semantic Graph, provide the robot as an information reference, and output low-dimensional actions. \citep{kazemi2020optimistic} uses the relevance between objects to generate Semantic Graph as prior knowledge to improve the success rate of task navigation.
Semantic Graph is represented by a graph, and the graph is extracted to obtain feature information. The directed graph $G=(V,E)$, $V$ is the node (Node), which represents the object in the image, and $E$ is the edge (Edge), which represents the correlation between the objects. There are many ways to obtain graph embedding by Graph Neural Network (GNN). GCN \citep{kipf2016semi} calculates the basic graph embedding; Graph Attention Networks \citep{velivckovic2017graph} using the attention mechanism to generate node weights; Graph Transformer Networks \citep{yun2019graph} generates more nodes and edges predict the new relationship between them to obtain a better graph embedding; Heterogeneous Graph Neural Network \citep{zhang2019heterogeneous} uses a variety of nodes and edges to generate more complex semantics.

\begin{figure}[b]
  \centering
  \includegraphics[width=0.7\linewidth]{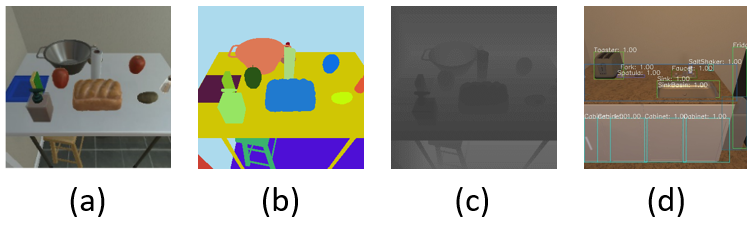}
  \caption{(a) Original image (b) Semantic segmentation (c) Depth image (d) Scene graph generation The result of recognizing the semantic segmentation image.}
  \label{img_explain}
\end{figure}

\section{Methods}
\label{headings}

The visual image and task instructions of the agent must be encoded, and then the agent outputs action operations. We use SGG to extract image features. SGG detects the current input image, recognizes the objects, extracts the relation between the object, object attribute, and gives the robot additional information for visual semantic processing.

\begin{figure}[t]
  \centering
  \includegraphics[width=0.99\linewidth]{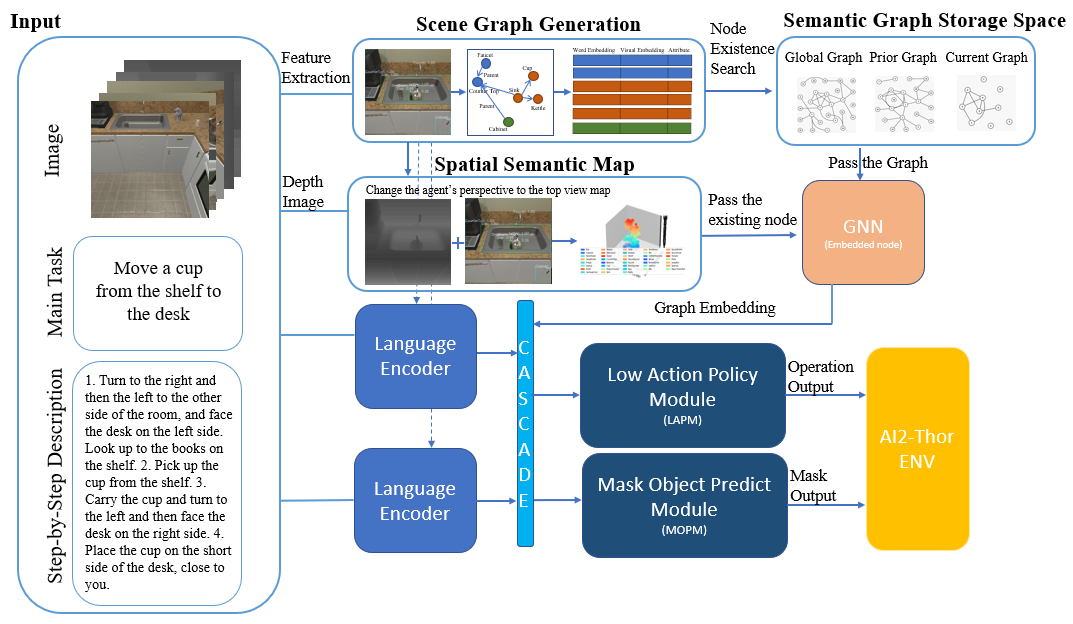}
  \caption{VSGM Details. After embedding and concatenating through input, SGG, language coding, and GNN, the actions are output through MOPM and LAPM.}
  \label{model}
\end{figure}

Figure~\ref{model} shows the basic architecture of our model. We use the MOCA \citep{singh2020moca} model as the benchmark model. Because the MOCA model directly uses the pre-trained model Resnet when encoding visual images, we believe that this approach is only suitable for object classification and cannot show the spatial orientation. Therefore, we input the RGB image $I^{rgb} \in \mathbb{R}^{h \text{x} w \text{x} 3}$ (Figure~\ref{img_explain}-a) into the SGG, and obtain the result (Figure~\ref{img_explain}-d) to generate the semantic graph. The spatial semantic map is the result obtained through SGG and the depth image $I^{d} \in \mathbb{R}^{h \text{x} w \text{x} 1}$ (Figure~\ref{img_explain}-c), which is mapped to the bird's-eye map. Later, The semantic graph and spatial semantic map use GNN to extract features of the graph, connect it to the middle part of the MOCA model, and then outputs the operation action. According to Figure~\ref{model}, after the SGG processing is completed, it will connect the visual embedding, and instruction embedding, perform the output of the low motion operation module (LAPM) and the interactive object module (MOPM). The output of LAPM is the value of 13 dimensions and the output of MOPM is the value of 119 dimensions, and the maximum value is input into the environment.

, where $t$ is
denotes
\subsection{Model Input}

The agent has 3 inputs, including the current image (Image), the main task (Main Task), and the step-by-step description (Step-by-step Description).

Visual coding. The agent can obtain the image information of the current state $I_t^{rgb},I_t^{d}$, $t$ is expressed as a time unit. $SGG(img)$ is the neural network of SGG, pre-trained from the ALFRED dataset. $O_{feat^{V}}^{c},r,A \gets SGG(I_t^{rgb})$, $O$ is image object, $feat^{V} \in \mathbb{R}^{1 \text{x} 2048}$ is the visual embedding of the object, $c \in \{0, \dots , 106\}$ is the object type, the object relevance $r \in \mathbb{R}^2$, and the object attribute $A$ $Attribute \in \mathbb{R}^{1 \text{x} 23}$.

Language encoding. The main task and detailed operation instructions are two inputs, each of which is coded through LSTM for language instructions. The language encoding method here is consistent with MOCA.

\subsection{Semantic Graph}

To extract higher-level features, it is necessary to consolidate the object information generated by SGG. $O_{feat^{V}}^{c},r,A$ establish the semantic graph $G^S = (V,E)$, $V$ represents the object $O$ in the image, and $E$ represents the object relevance $r$. The node feature of $V$ is 
\begin{equation}
feat^N = feat^{V} || A || feat^{word}
\label{embed_feature}
\end{equation}
where $feat^{word}$ is the word embedding obtained by \citep{bojanowski2017enriching}, and $||$ is the concatenated vector. $\forall V$ features in $G^S$ can express as $feat^{SG} \in \mathbb{R}^{N \text{x} 2371}$. $E$ represents the adjacency matrix $Adj \in \mathbb{R}^{N \text{x} N}$, and $N$ is the detected object.

There are three types of Semantic Graphs.

\textbf{Semantic Prior Graph} $G^{SP}$ establishes the relationship between all objects in the simulated environment and allows the agent to determine the relationship between the objects. The adjacency matrix is based on the object relationship provided by the Visual Genome dataset \citep{krishna2017visual}. Semantic Prior Graph is generated in advance and will not update during task execution.

\begin{algorithm}[t]
\SetAlgoLined
\DontPrintSemicolon
\SetKwInput{KwInput}{Input}
 \KwInput{$O, r, A \gets SGG(I_t^{rgb})$}
  \tcp*[l]{Initialize}
  $G_t^{SC} \in G^S = (V,E)$\;
  \For{each $O, A$}    
  {
    Get object visual feature $feat^{V}$, object name by $O_{feat^{V}}^{c}$, and object word embedding $feat^{word}$ by object name \;
    $feat^N = feat^{V} || A || feat^{word}$ \tcp*[r]{Set node feature, Eq.~\ref{embed_feature}}
    $G_t^{SC}$ adds a new node with node feature $feat^N$ \;
  }
  \For{each $r$}    
  {
    src, dst $ \gets r$ \;
    Update $G_t^{SC} Adj:$ $dst \gets src$ \;
  }
 \KwRet $G_t^{SC}$ \;
 \caption{Update Semantic Current Graph}
 \label{Semantic_Current_Graph}
\end{algorithm}
\begin{algorithm}[t]
\SetAlgoLined
\DontPrintSemicolon
\SetKwInput{KwInput}{Input}
 \KwInput{$G_{t - 1}^{SG}, G_{t}^{SC}$}
  \tcp*[l]{Initialize}
  \If{t == 0} 
  {
      Set $Threshold$ to a constant value \tcp*[r]{$Threshold \in [0,1]$}
      $G_{0}^{SG} \in G^S = (V,E)$ \;
  }
  $G_{t}^{SG} \gets G_{t - 1}^{SG}$ \;
  \For{each $V, E$ in $G_{t}^{SC}$ }
  {
    Use cosine similarity to calculate the similarity between $V$ 's node feature and $\forall V^{'}$ 's node feature \tcp*[r]{ $ V^{'} \in G_{t - 1}^{SG} $ }
    \If{similarity < $Threshold$}
    {
      $G_{t}^{SG}$ adds a new node and edge by $V, E$ \;
    }
  }
  
 \KwRet $G_{t}^{SG}$ \;
 \caption{Update Semantic Global Graph}
 \label{Semantic_Global_Graph}
\end{algorithm}

\textbf{Semantic Current Graph} $G_t^{SC}$ represents the semantic graph of the current state and provides the agent to use this semantic graph as a reference. Whenever the status is updated, Semantic Current Graph will refresh again. Please refer to Algorithm~\ref{Semantic_Current_Graph} for the detailed process.

\textbf{Semantic Global Graph} $G_t^{SG}$ records the graph of the state of all objects so far, and the agent can use this as a reference to determine the completion of the task. Whenever the status is updated, Semantic Global Graph will be updated. $G_t^{SG} = Update(G_{t - 1}^{SG}, G_{t}^{SC})$, $G_{t - 1}^{SG}$ is the previous state. When an object is detected, to avoid a repeated accumulation of nodes, the same type of object that has appeared in $G_{t - 1}^{SG}$ will compare with the node feature of the two using cosine similarity. When the similarity is less than $Threshold$, it will be added to $G_t^{SG}$ as a new object to complete the update. For details, please refer to Algorithm~\ref{Semantic_Global_Graph}.

\subsection{Spatial Semantic Map}

In addition to Semantic Graph, the robot's movement in the environment, recording object positions will help navigation and planning. The object $O$ detected by SGG and the depth image $I_t^d$ will generate top-down egocentric Semantic Maps $M_t \in \mathbb{R}^{s \text{x} s \text{x} c}$, $s$ means the width of this map, $c = \{1, \dots ,106\}$ means the type of object. When there is an object at the $(i,j), \, i,j \in \{1, \dots , s\}$ space position, $M_t^{(i,j)}$ will be marked with this object.

The agent starts at the center of $(s/2,s/2)$. First, in the egocentric state of the robot's perspective, the pixel coordinates of the depth image $I_t^d$ correspond to the pixel coordinates of the object $O$ detected by SGG. $(x,y,c), \, \forall x \in \{1, \dots ,i\}, \forall y \in \{1, \dots ,j\}, c=\{1, \dots , 106\}$. $i$ and $j$ are the detected object ranges. After $I_t^d$ is converted into a point cloud, the agent's perspective transforms into a top-down perspective, which represents the agent's object position in the environment space. $M_t$ only represents the position of the object in the spatial environment. To allow the map to express the features of the object, $M_t$ is stored in a graph-based map, $G_t^M = (V, E), \, \forall V \in \{1, 2, \dots , N\}$, the embedding feature of $V$ is the same as $feat^N$.
The feature of all nodes $\forall V$ of $G_t^M$ is $feat^{M} \in \mathbb{R}^{N \text{x} 2371 }, N = s \text{x} s \text{x} c $, which has the same object feature, object attribute, and word embedding as the $V$ of $G^S$. Whenever the status is updated and new $I_t^{rgb}$, $I_t^d$ is obtained, the spatial semantic map will be updated $G_t^M = Update(G_{t - 1}^M, SGG(I_t^{rgb}), I_t^d)$, please refer to Algorithm~\ref{Semantic_Maps} for details.

\begin{algorithm}[t]
\SetAlgoLined
\DontPrintSemicolon
\SetKwInput{KwInput}{Input}
 \KwInput{ $ G_{t - 1}^M, I_t^d, O, r, A \gets SGG(I_t^{rgb}) $ }
  \tcp*[l]{Initialize}
  \If{t == 0} 
  {
      Set $S, C$ to a constant value \tcp*[r]{$S$ is the map size. $C$ is all objects that exist in the environment.}
      $G_{0}^{M} \in G^S = (V,E)$ \;
      $G_{0}^{M}$ adds $S \text{x} S \text{x} C$ nodes \;
      Connect the node $V^i$ and the node $V^j$ if the nodes are neighbors. \tcp*[r]{$\forall V^i, V^j \in G_{0}^{M}, i \neq j$} 
  }
  $G_{t}^{M} \gets G_{t - 1}^{M}$ \;
  \For{each O}
  {
    \For{$i, j \gets O $'s bounding box coordinate}
    {
      $depth \gets I_t^d (i, j) $ \tcp*[r]{The depth of the agent's perspective of the point cloud}
      $(u, v) \gets depth \cdot M^T$ \tcp*[r]{$M^T$ is a matrix transform the agent's perspective to top-down view from the top}
      Get object visual feature $feat^{V}$, object name by $O_{feat^{V}}^{c}$, and object word embedding $feat^{word}$ by object name \;
      $feat^N = feat^{V} || A || feat^{word}$ \tcp*[r]{Set node feature, Eq.~\ref{embed_feature}}
      $G_{t}^{M}(u,v,c) \gets feat^N$ \tcp*[r]{Set the current space location to exist this object}
    }
  }
 \KwRet $G_{t}^{M}$ \;
 \caption{Update Semantic Maps}
 \label{Semantic_Maps}
\end{algorithm}

\subsection{Graph Neural Networks}

After the semantic graph and spatial semantic map are updated, GNN is used to extract vital features to obtain a high-dimensional representation vector.
Take GCN \citep{kipf2016semi} as an example, $GCN(G), G \in \{ G^{SP}, G_t^{SC}, G_t^{SG}, G_t^{M} \}$. When $G_t^M$ performs graph embedding, only nodes $V$ with objects in $M_t^{(i,j)}$ are selected, and the rest are ignored. The following is the information transmission form of GCN
\begin{equation}
h^i = softmax( \hat{A} \cdot \sigma (\hat{A} \cdot h^{i-1} \cdot W_1^{i - 1}) \cdot W_2^{i - 1} )
\label{gnn}
\end{equation}
where $h^i$ represents the eigenvalue of the node after the output of the neural network, $\hat{A}$ represents the adjacency matrix, $\alpha$ represents the activation function, $h^{i - 1}$ represents the feature of the node at the previous moment, $W$ represents the network weight. With the graph neural networks, the embedding of the graph can be obtained.

After obtaining the $h^i$ of the final layer, use $GraphEmbed(h)$ to embed the nodes in a graph, as shown below
\begin{equation}
\begin{split}
& X^{GE} = GraphEmbed(h) = \sigma ( \sum_{j=1}^N (\alpha \cdot W_1 \cdot h_{V^j}^i) ) \\
& \alpha = softmax(e^j) = \frac{exp(e^j)}{\sum_{k=1}^{N} exp(e^k)}  \\
& e^j = a( W_2 \cdot h_{V^j}^i, (W_3^ \cdot H_t^{lang})^T )
\label{graph_embed}
\end{split}
\end{equation}
where $\alpha$ is weight coefficient, $e^j$ is attention coefficients, $a$ could be an inner product, and $H_t^{lang}$ is the hidden state of the language model LSTM. As the agent performs tasks, $H_t^{lang}$ will also keep changing.

\subsection{Objective Function}

The following is the objective function when we train the model
\begin{equation}
L = - \sum_{t = 0}^{T} y_t^a \cdot \log_{}{p_t^a} \, - \sum_{t = 0}^{T} y_t^c \cdot \log_{}{p_t^c}
\end{equation}
where $y_t^a$ and $y_t^c$ respectively represent the action displayed by the expert (Ground true label), and $p_t^a$ and $p_t^c$ respectively represent the probability of the model's low motion operation module (LAPM) and interactive object module (MOPM) predicting the output. $T$ represents the total expert operation trajectory (Trajectory), $t$ represents the action selection at the current state. The model will output actions through LAPM and MOPM at every moment.

\section{Experiments}
\label{others}

In this paper, we use the ALFRED dataset as the criterion for our experimental evaluation. ALFRED is a complete daily life task-oriented data collection, based on AI2-THOR \citep{kolve2017ai2} as the simulation environment, to control the robot to complete the task of language instruction. The dataset contains different types of main tasks and basic action operations (See Supplementary Material for details). The daily life environment includes 120 kinds of living rooms, bedrooms, kitchens, and bathrooms. ALFRED can evaluate the semantic map by giving the final task whether it succeeded or failed. We want to know whether the proposed semantic method can effectively help strategy, inference, and better interpretability. 

\textbf{Tasks description}. Contains a rough description of the tasks to be completed, and detailed operations (Step-by-step) of the agent following language instructions. Mainly divided into 7 main task types.

\textbf{Basic operations}. The most basic operations that the agent can perform. When Stop is executed, the task is stopped. Navigation and perspective operations include LookDown, LookUp, MoveAhead, RotateLeft, RotateRight, and operations on scene objects include PickupObject, SliceObject, OpenObject, PutObject, CloseObject, ToggleObjectOn, ToggleObjectOff. In addition to operation actions, object operations must also include The mask of the operating position in the simulated environment.

\begin{table}[t]
  \caption{Main task type. Seen and Unseen mean seen the scene during training. Task means task success rate, GC mean Goal-Cond. $\bullet$ means segmentation and depth image; $*$ means using SGG model. We select the best value after training the model with two random seed.}
  \label{main_task}
  \centering
  \begin{tabular}{p{0.3\linewidth}  p{0.1\linewidth}  p{0.1\linewidth}  p{0.1\linewidth}  p{0.1\linewidth}  p{0.1\linewidth}}
    \toprule
    Model &Validation && \multicolumn{3}{l}{Sub-Goal Validation} \\
          &(\%)       && (\%) \\
    \midrule
    & Seen & Unseen & Seen\,(Unseen)\\
    \midrule
    & Task\,(GC) & Task\,(GC) & Goto & Pickup & Put \\
    \midrule
    Seq2Seq+PM \citep{shridhar2020alfred} & 0\,\,\,(0) & 0\,(0) & 20\,(34) & 31\,(21) & \textbf{64\,(11)}     \\
    MOCA \citep{singh2020moca} &  5\,\,\,(5) & 0\,(0) & 40\,(42) & 65\,(42) & 6\,\,\,\,(0)      \\
    MOCA $\bullet$  & 3\,\,\,(3) & 0\,(0) & 39\,(43) & 64\,(53) & 7\,\,\,\,(6) \\
    Oracle Semantic Graph  & 6\,\,\,(6) & 0\,(0) & 46\,(39) & 66\,(37) & 7\,\,\,\,(5)\\
    Oracle Semantic Graph $\bullet$  & 8\,\,\,(8) & 0\,(0) & 44\,(48) & 67\,(48) & 5\,\,\,\,(10) \\
    Semantic Graph $\bullet$*  & 12\,(12) & 0\,(0) & 47\,(46) & \textbf{69\,(63)} & 14\,\,(4) \\
    Semantic Graph $\bullet$* + SLAM  & 8\,\,\,(8) & 0\,(0) & \textbf{55\,(55)} & 47\,(40) & 10\,\,(7)  \\
    VSGM $\bullet$* & \textbf{15\,(15)} & \textbf{1\,(1)} & 54\,(47) & 60\,(46) & 16\,\,(0) \\
    \midrule
    HUMAN  & 91.0 & 94.5 & & &  \\
    \bottomrule
  \end{tabular}
\end{table}

\subsection{Metrics}
\paragraph{Evaluation Metrics} There are two ways to evaluate, Task Success and Sub-Goal success rate. If the object's position or state changes and the Goal-Condition is reached, the main task success (Task Success) is defined as 1; otherwise, it is 0. The second evaluation method is Sub-Goal Evaluation. For the main task (Main Task), there may be several sub-tasks (Sub-Goals) combined, including navigation (GoTo), pick up (Pickup), put down (Put), cooling (Cool), heating (Heat), Cleaning (Clean), cutting (Slice), switching (Toggle), to form the main task. The environment can individually assess whether its subtasks meet the task conditions. The Path weighted success rate compares the length difference between the agent and the expert's display path. For details, please refer to ALFRED.

\subsection{Result}

\textbf{Experimental data}. All experimental data are placed in Table~\ref{main_task}. We conduct experiments on picking and placing. Oracle's model building method is the ground true environment data provided by AI2-THOR, which includes visual objects, object attributes, and object correlations to establish a semantic graph. Regardless of whether it is Oracle, after adding semantic assistance, the model can have the effect of improving the success rate of the task. VSGM is a combination of SGG, Semantic Graph, and Spatial Semantic Maps.

\textbf{Semantic Graph} is used to allow the robot to extract important nodes, allowing more attention to key objects. It can be more effective in the scene task ALFRED dataset, and achieve better results. And we also tried to modify the current input image in this research, replacing the RGB image with a semantic segmentation image $I_t^{seg} \in \mathbb{R}^{h \text{x} w \text{x} 3}$ (Figure~\ref{img_explain}-c) and a depth image, to observe the impact on task execution. It show that the input of semantic segmentation image and depth image improves the model performance.

\begin{figure}[t]
  \centering
  \includegraphics[width=0.7\linewidth]{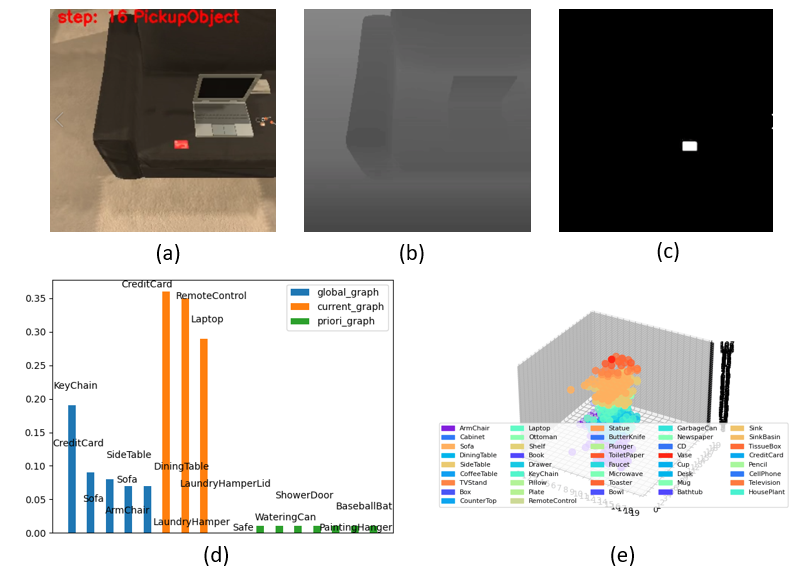}
  \caption{Show a successful case. (a) is the current input image, the upper left corner shows that the LAPM action output by the model is to pick up the object; (b) is the current depth image; (c) is the mask that MOPM expects to pick up the object from the model output; (d) Attention score for Semantic Graph. The blue bar represents Semantic Global Graph, the orange bar represents Semantic Current Graph, and the green bar represents Semantic Prior Graph; (e) is Spatial Semantic Map.}
  \label{sf6}
\end{figure}

\textbf{Semantic Map performance}. To verify the semantic map method we proposed, we combined the SLAM of \citep{chaplot2020learning} with the Semantic graph to observe the results. We believe that the SLAM method can focus on navigation and exploration, but the ALFRED task is the subtle operation of the robot. If only the information of space and obstacles is given to the robot, the prediction effect will not be very well. Table~\ref{main_task} shows that as a result of adding SLAM, the navigation ability of Sub-Goals has been improved, but the success rate of the VSGM task is better.

Figure~\ref{sf6} can see that the model has more attention weight in "CreditCard" of the current language intention, which shows that the model considers it has a more important influence. However, the attention weight of a priori language intention is more scattered. We believe that due to the limitation of training data, the task is too complex, and the objects of prior knowledge are too many, the model cannot be trained well.

\subsection{Ablation Study and Other Experiments}

We ablate the three types of Semantic Graph. Use Sub-Goal success rate to evaluate. In summary, the combination of three Semantic Graphs has a better effect on the results. There is no significant difference between assigning different Semantic Graphs to LAPM and MOPM. Semantic Graphs can be directly connected. When Semantic Graph is added to RGB and MaskDepth, the MaskDepth effect will be better than RGB. No matter what kind of Semantic Graph is concatenated, the model results can be improved. Please refer to Supplementary for more experimental data.

\section{Conclusion}

In this paper, we propose Semantic Graph and Spatial Semantic Map, using nodes to explore whether the neural network can improve the agent's judgment ability on the characteristics of objects. We have shown that that VSGM can effectively improve the agent's ability. In the future, our research can add the next goal prediction of the semantic map, the semantic intent with multi-layer meta-path, and more helpful information about attributes and relations. We study whether the features of the objects represented by the graph can increase the understanding of the robot. There is still a lot of room for optimization for the understanding of language instructions.

\bibliographystyle{plain}
\medskip

{
\small

\bibliography{references}

\begin{thebibliography}{10}

\bibitem{anderson2018vision}
Peter Anderson, Qi~Wu, Damien Teney, Jake Bruce, Mark Johnson, Niko
  S{\"u}nderhauf, Ian Reid, Stephen Gould, and Anton Van Den~Hengel.
\newblock Vision-and-language navigation: Interpreting visually-grounded
  navigation instructions in real environments.
\newblock In {\em Proceedings of the IEEE Conference on Computer Vision and
  Pattern Recognition}, pages 3674--3683, 2018.

\bibitem{armeni20193d}
Iro Armeni, Zhi-Yang He, JunYoung Gwak, Amir~R Zamir, Martin Fischer, Jitendra
  Malik, and Silvio Savarese.
\newblock 3d scene graph: A structure for unified semantics, 3d space, and
  camera.
\newblock In {\em Proceedings of the IEEE/CVF International Conference on
  Computer Vision}, pages 5664--5673, 2019.

\bibitem{beeching2020learning}
Edward Beeching, Jilles Dibangoye, Olivier Simonin, and Christian Wolf.
\newblock Learning to plan with uncertain topological maps.
\newblock {\em arXiv preprint arXiv:2007.05270}, 2020.

\bibitem{bojanowski2017enriching}
Piotr Bojanowski, Edouard Grave, Armand Joulin, and Tomas Mikolov.
\newblock Enriching word vectors with subword information.
\newblock {\em Transactions of the Association for Computational Linguistics},
  5:135--146, 2017.

\bibitem{cartillier2020semantic}
Vincent Cartillier, Zhile Ren, Neha Jain, Stefan Lee, Irfan Essa, and Dhruv
  Batra.
\newblock Semantic mapnet: Building allocentric semanticmaps and
  representations from egocentric views.
\newblock {\em AAAI}, 2021.

\bibitem{chang2017matterport3d}
Angel Chang, Angela Dai, Thomas Funkhouser, Maciej Halber, Matthias Niessner,
  Manolis Savva, Shuran Song, Andy Zeng, and Yinda Zhang.
\newblock Matterport3d: Learning from rgb-d data in indoor environments.
\newblock {\em arXiv preprint arXiv:1709.06158}, 2017.

\bibitem{chaplot2020learning}
Devendra~Singh Chaplot, Dhiraj Gandhi, Saurabh Gupta, Abhinav Gupta, and Ruslan
  Salakhutdinov.
\newblock Learning to explore using active neural slam.
\newblock {\em arXiv preprint arXiv:2004.05155}, 2020.

\bibitem{chaplot2020neural}
Devendra~Singh Chaplot, Ruslan Salakhutdinov, Abhinav Gupta, and Saurabh Gupta.
\newblock Neural topological slam for visual navigation.
\newblock In {\em Proceedings of the IEEE/CVF Conference on Computer Vision and
  Pattern Recognition}, pages 12875--12884, 2020.

\bibitem{chen2020learning}
Changan Chen, Sagnik Majumder, Ziad Al-Halah, Ruohan Gao, Santhosh~Kumar
  Ramakrishnan, and Kristen Grauman.
\newblock Learning to set waypoints for audio-visual navigation.
\newblock {\em ICLR}, 2021.

\bibitem{chen2020topological}
Kevin Chen, Junshen~K Chen, Jo~Chuang, Marynel V{\'a}zquez, and Silvio
  Savarese.
\newblock Topological planning with transformers for vision-and-language
  navigation.
\newblock {\em arXiv preprint arXiv:2012.05292}, 2020.

\bibitem{du2020learning}
Heming Du, Xin Yu, and Liang Zheng.
\newblock Learning object relation graph and tentative policy for visual
  navigation.
\newblock In {\em European Conference on Computer Vision}, pages 19--34.
  Springer, 2020.

\bibitem{fried2018speaker}
Daniel Fried, Ronghang Hu, Volkan Cirik, Anna Rohrbach, Jacob Andreas,
  Louis-Philippe Morency, Taylor Berg-Kirkpatrick, Kate Saenko, Dan Klein, and
  Trevor Darrell.
\newblock Speaker-follower models for vision-and-language navigation.
\newblock {\em NIPS}, 2018.

\bibitem{gan2020look}
Chuang Gan, Yiwei Zhang, Jiajun Wu, Boqing Gong, and Joshua~B Tenenbaum.
\newblock Look, listen, and act: Towards audio-visual embodied navigation.
\newblock In {\em 2020 IEEE International Conference on Robotics and Automation
  (ICRA)}, pages 9701--9707. IEEE, 2020.

\bibitem{gordon2019splitnet}
Daniel Gordon, Abhishek Kadian, Devi Parikh, Judy Hoffman, and Dhruv Batra.
\newblock Splitnet: Sim2sim and task2task transfer for embodied visual
  navigation.
\newblock In {\em Proceedings of the IEEE/CVF International Conference on
  Computer Vision}, pages 1022--1031, 2019.

\bibitem{he2017mask}
Kaiming He, Georgia Gkioxari, Piotr Doll{\'a}r, and Ross Girshick.
\newblock Mask r-cnn.
\newblock In {\em Proceedings of the IEEE international conference on computer
  vision}, pages 2961--2969, 2017.

\bibitem{he2016deep}
Kaiming He, Xiangyu Zhang, Shaoqing Ren, and Jian Sun.
\newblock Deep residual learning for image recognition.
\newblock In {\em Proceedings of the IEEE conference on computer vision and
  pattern recognition}, pages 770--778, 2016.

\bibitem{jain2020cordial}
Unnat Jain, Luca Weihs, Eric Kolve, Ali Farhadi, Svetlana Lazebnik, Aniruddha
  Kembhavi, and Alexander Schwing.
\newblock A cordial sync: Going beyond marginal policies for multi-agent
  embodied tasks.
\newblock In {\em European Conference on Computer Vision}, pages 471--490.
  Springer, 2020.

\bibitem{kazemi2020optimistic}
Mahdi Kazemi~Moghaddam, Qi~Wu, Ehsan Abbasnejad, and Javen Qinfeng~Shi.
\newblock Optimistic agent: Accurate graph-based value estimation for more
  successful visual navigation.
\newblock {\em WACV}, 2021.

\bibitem{kim20193}
Ue-Hwan Kim, Jin-Man Park, Taek-Jin Song, and Jong-Hwan Kim.
\newblock 3-d scene graph: A sparse and semantic representation of physical
  environments for intelligent agents.
\newblock {\em IEEE transactions on cybernetics}, 50(12):4921--4933, 2019.

\bibitem{kipf2016semi}
Thomas~N Kipf and Max Welling.
\newblock Semi-supervised classification with graph convolutional networks.
\newblock {\em ICLR}, 2017.

\bibitem{kolve2017ai2}
Eric Kolve, Roozbeh Mottaghi, Winson Han, Eli VanderBilt, Luca Weihs, Alvaro
  Herrasti, Daniel Gordon, Yuke Zhu, Abhinav Gupta, and Ali Farhadi.
\newblock Ai2-thor: An interactive 3d environment for visual ai.
\newblock {\em arXiv preprint arXiv:1712.05474}, 2017.

\bibitem{krantz2020beyond}
Jacob Krantz, Erik Wijmans, Arjun Majumdar, Dhruv Batra, and Stefan Lee.
\newblock Beyond the nav-graph: Vision-and-language navigation in continuous
  environments.
\newblock In {\em European Conference on Computer Vision}, pages 104--120.
  Springer, 2020.

\bibitem{krishna2017visual}
Ranjay Krishna, Yuke Zhu, Oliver Groth, Justin Johnson, Kenji Hata, Joshua
  Kravitz, Stephanie Chen, Yannis Kalantidis, Li-Jia Li, David~A Shamma, et~al.
\newblock Visual genome: Connecting language and vision using crowdsourced
  dense image annotations.
\newblock {\em International journal of computer vision}, 123(1):32--73, 2017.

\bibitem{li2018factorizable}
Yikang Li, Wanli Ouyang, Bolei Zhou, Jianping Shi, Chao Zhang, and Xiaogang
  Wang.
\newblock Factorizable net: an efficient subgraph-based framework for scene
  graph generation.
\newblock In {\em Proceedings of the European Conference on Computer Vision
  (ECCV)}, pages 335--351, 2018.

\bibitem{liao2020tsm}
Zhiyong Liao, Yu~Zhang, Junren Luo, and Weilin Yuan.
\newblock Tsm: Topological scene map for representation in indoor environment
  understanding.
\newblock {\em IEEE Access}, 8:185870--185884, 2020.

\bibitem{lv2020improving}
Yunlian Lv, Ning Xie, Yimin Shi, Zijiao Wang, and Heng~Tao Shen.
\newblock Improving target-driven visual navigation with attention on 3d
  spatial relationships.
\newblock {\em arXiv preprint arXiv:2005.02153}, 2020.

\bibitem{majumdar2020improving}
Arjun Majumdar, Ayush Shrivastava, Stefan Lee, Peter Anderson, Devi Parikh, and
  Dhruv Batra.
\newblock Improving vision-and-language navigation with image-text pairs from
  the web.
\newblock In {\em European Conference on Computer Vision}, pages 259--274.
  Springer, 2020.

\bibitem{misra2018mapping}
Dipendra Misra, Andrew Bennett, Valts Blukis, Eyvind Niklasson, Max Shatkhin,
  and Yoav Artzi.
\newblock Mapping instructions to actions in 3d environments with visual goal
  prediction.
\newblock {\em EMNLP}, 2018.

\bibitem{nguyen2019reinforcement}
Tai-Long Nguyen, Do-Van Nguyen, and Thanh-Ha Le.
\newblock Reinforcement learning based navigation with semantic knowledge of
  indoor environments.
\newblock In {\em 2019 11th International Conference on Knowledge and Systems
  Engineering (KSE)}, pages 1--7. IEEE, 2019.

\bibitem{qiulearning}
Yiding Qiu, Anwesan Pal, and Henrik~I Christensen.
\newblock Learning hierarchical relationships for object-goal navigation.
\newblock {\em CoRL}, 2020.

\bibitem{redmon2016you}
Joseph Redmon, Santosh Divvala, Ross Girshick, and Ali Farhadi.
\newblock You only look once: Unified, real-time object detection.
\newblock In {\em Proceedings of the IEEE conference on computer vision and
  pattern recognition}, pages 779--788, 2016.

\bibitem{rosinol20203d}
Antoni Rosinol, Arjun Gupta, Marcus Abate, Jingnan Shi, and Luca Carlone.
\newblock 3d dynamic scene graphs: Actionable spatial perception with places,
  objects, and humans.
\newblock {\em arXiv preprint arXiv:2002.06289}, 2020.

\bibitem{saha2021modular}
Homagni Saha, Fateme Fotouhif, Qisai Liu, and Soumik Sarkar.
\newblock A modular vision language navigation and manipulation framework for
  long horizon compositional tasks in indoor environment.
\newblock {\em arXiv preprint arXiv:2101.07891}, 2021.

\bibitem{savva2019habitat}
Manolis Savva, Abhishek Kadian, Oleksandr Maksymets, Yili Zhao, Erik Wijmans,
  Bhavana Jain, Julian Straub, Jia Liu, Vladlen Koltun, Jitendra Malik, et~al.
\newblock Habitat: A platform for embodied ai research.
\newblock 2019.

\bibitem{seymour2021maast}
Zachary Seymour, Kowshik Thopalli, Niluthpol Mithun, Han-Pang Chiu, Supun
  Samarasekera, and Rakesh Kumar.
\newblock Maast: Map attention with semantic transformersfor efficient visual
  navigation.
\newblock {\em ICRA}, 2021.

\bibitem{shen2021spatial}
Zhengcheng Shen, Linh K{\"a}stner, and Jens Lambrecht.
\newblock Spatial imagination with semantic cognition for mobile robots.
\newblock {\em arXiv preprint arXiv:2104.03638}, 2021.

\bibitem{shridhar2020alfred}
Mohit Shridhar, Jesse Thomason, Daniel Gordon, Yonatan Bisk, Winson Han,
  Roozbeh Mottaghi, Luke Zettlemoyer, and Dieter Fox.
\newblock Alfred: A benchmark for interpreting grounded instructions for
  everyday tasks.
\newblock In {\em Proceedings of the IEEE/CVF conference on computer vision and
  pattern recognition}, pages 10740--10749, 2020.

\bibitem{shridhar2020alfworld}
Mohit Shridhar, Xingdi Yuan, Marc-Alexandre C{\^o}t{\'e}, Yonatan Bisk, Adam
  Trischler, and Matthew Hausknecht.
\newblock Alfworld: Aligning text and embodied environments for interactive
  learning.
\newblock {\em ICLR}, 2021.

\bibitem{singh2020moca}
Kunal~Pratap Singh, Suvaansh Bhambri, Byeonghwi Kim, Roozbeh Mottaghi, and
  Jonghyun Choi.
\newblock Moca: A modular object-centric approach for interactive instruction
  following.
\newblock {\em arXiv preprint arXiv:2012.03208}, 2020.

\bibitem{storks2021we}
Shane Storks, Qiaozi Gao, Govind Thattai, and Gokhan Tur.
\newblock Are we there yet? learning to localize in embodied instruction
  following.
\newblock {\em AAAI}, 2021.

\bibitem{tan2019learning}
Hao Tan, Licheng Yu, and Mohit Bansal.
\newblock Learning to navigate unseen environments: Back translation with
  environmental dropout.
\newblock {\em NAACL}, 2019.

\bibitem{tang2020unbiased}
Kaihua Tang, Yulei Niu, Jianqiang Huang, Jiaxin Shi, and Hanwang Zhang.
\newblock Unbiased scene graph generation from biased training.
\newblock In {\em Proceedings of the IEEE/CVF Conference on Computer Vision and
  Pattern Recognition}, pages 3716--3725, 2020.

\bibitem{thomason2020vision}
Jesse Thomason, Michael Murray, Maya Cakmak, and Luke Zettlemoyer.
\newblock Vision-and-dialog navigation.
\newblock In {\em Conference on Robot Learning}, pages 394--406. PMLR, 2020.

\bibitem{velivckovic2017graph}
Petar Veli{\v{c}}kovi{\'c}, Guillem Cucurull, Arantxa Casanova, Adriana Romero,
  Pietro Lio, and Yoshua Bengio.
\newblock Graph attention networks.
\newblock {\em ICLR}, 2018.

\bibitem{wang2019reinforced}
Xin Wang, Qiuyuan Huang, Asli Celikyilmaz, Jianfeng Gao, Dinghan Shen,
  Yuan-Fang Wang, William~Yang Wang, and Lei Zhang.
\newblock Reinforced cross-modal matching and self-supervised imitation
  learning for vision-language navigation.
\newblock In {\em Proceedings of the IEEE/CVF Conference on Computer Vision and
  Pattern Recognition}, pages 6629--6638, 2019.

\bibitem{wang2020environment}
Xin Wang, Vihan Jain, Eugene Ie, William~Yang Wang, Zornitsa Kozareva, and
  Sujith Ravi.
\newblock Environment-agnostic multitask learning for natural language grounded
  navigation.
\newblock {\em ECCV}, 2020.

\bibitem{wani2020multion}
Saim Wani, Shivansh Patel, Unnat Jain, Angel~X Chang, and Manolis Savva.
\newblock Multion: Benchmarking semantic map memory using multi-object
  navigation.
\newblock {\em NIPS}, 2020.

\bibitem{wijmans2019dd}
Erik Wijmans, Abhishek Kadian, Ari Morcos, Stefan Lee, Irfan Essa, Devi Parikh,
  Manolis Savva, and Dhruv Batra.
\newblock Dd-ppo: Learning near-perfect pointgoal navigators from 2.5 billion
  frames.
\newblock {\em ICLR}, 2020.

\bibitem{wortsman2019learning}
Mitchell Wortsman, Kiana Ehsani, Mohammad Rastegari, Ali Farhadi, and Roozbeh
  Mottaghi.
\newblock Learning to learn how to learn: Self-adaptive visual navigation using
  meta-learning.
\newblock In {\em Proceedings of the IEEE/CVF Conference on Computer Vision and
  Pattern Recognition}, pages 6750--6759, 2019.

\bibitem{xu2019regression}
Danfei Xu, Roberto Mart{\'\i}n-Mart{\'\i}n, De-An Huang, Yuke Zhu, Silvio
  Savarese, and Li~Fei-Fei.
\newblock Regression planning networks.
\newblock {\em NeurIPS}, 2019.

\bibitem{yang2018graph}
Jianwei Yang, Jiasen Lu, Stefan Lee, Dhruv Batra, and Devi Parikh.
\newblock Graph r-cnn for scene graph generation.
\newblock In {\em Proceedings of the European conference on computer vision
  (ECCV)}, pages 670--685, 2018.

\bibitem{yang2018visual}
Wei Yang, Xiaolong Wang, Ali Farhadi, Abhinav Gupta, and Roozbeh Mottaghi.
\newblock Visual semantic navigation using scene priors.
\newblock {\em arXiv preprint arXiv:1810.06543}, 2018.

\bibitem{yun2019graph}
Seongjun Yun, Minbyul Jeong, Raehyun Kim, Jaewoo Kang, and Hyunwoo~J Kim.
\newblock Graph transformer networks.
\newblock {\em NeurIPS}, 2019.

\bibitem{zhang2019heterogeneous}
Chuxu Zhang, Dongjin Song, Chao Huang, Ananthram Swami, and Nitesh~V Chawla.
\newblock Heterogeneous graph neural network.
\newblock In {\em Proceedings of the 25th ACM SIGKDD International Conference
  on Knowledge Discovery \& Data Mining}, pages 793--803, 2019.

\bibitem{zhu2020hierarchical}
Yifeng Zhu, Jonathan Tremblay, Stan Birchfield, and Yuke Zhu.
\newblock Hierarchical planning for long-horizon manipulation with geometric
  and symbolic scene graphs.
\newblock {\em ICRA}, 2021.

\bibitem{zhu2017visual}
Yuke Zhu, Daniel Gordon, Eric Kolve, Dieter Fox, Li~Fei-Fei, Abhinav Gupta,
  Roozbeh Mottaghi, and Ali Farhadi.
\newblock Visual semantic planning using deep successor representations.
\newblock In {\em Proceedings of the IEEE international conference on computer
  vision}, pages 483--492, 2017.

\end{thebibliography}
}

\section*{Checklist}

\begin{enumerate}

\item For all authors...
\begin{enumerate}
  \item Do the main claims made in the abstract and introduction accurately reflect the paper's contributions and scope?
    \answerYes{}
  \item Did you describe the limitations of your work?
    \answerYes{}
  \item Did you discuss any potential negative societal impacts of your work?
    \answerNo{}
  \item Have you read the ethics review guidelines and ensured that your paper conforms to them?
    \answerYes{}
\end{enumerate}

\item If you are including theoretical results...
\begin{enumerate}
  \item Did you state the full set of assumptions of all theoretical results?
    \answerNA{Yes}
  \item Did you include complete proofs of all theoretical results?
    \answerNA{}
\end{enumerate}

\item If you ran experiments...
\begin{enumerate}
  \item Did you include the code, data, and instructions needed to reproduce the main experimental results (either in the supplemental material or as a URL)?
    \answerYes{}
  \item Did you specify all the training details (e.g., data splits, hyperparameters, how they were chosen)?
    \answerYes{}
  \item Did you report error bars (e.g., with respect to the random seed after running experiments multiple times)?
    \answerNo{}
  \item Did you include the total amount of compute and the type of resources used (e.g., type of GPUs, internal cluster, or cloud provider)?
    \answerYes{}
\end{enumerate}

\item If you are using existing assets (e.g., code, data, models) or curating/releasing new assets...
\begin{enumerate}
  \item If your work uses existing assets, did you cite the creators?
    \answerYes{}
  \item Did you mention the license of the assets?
    \answerNo{}
  \item Did you include any new assets either in the supplemental material or as a URL?
    \answerNA{}
  \item Did you discuss whether and how consent was obtained from people whose data you're using/curating?
    \answerNA{The ALFRED dataset can be downloaded on Github}
  \item Did you discuss whether the data you are using/curating contains personally identifiable information or offensive content?
    \answerNo{}
\end{enumerate}

\item If you used crowdsourcing or conducted research with human subjects...
\begin{enumerate}
  \item Did you include the full text of instructions given to participants and screenshots, if applicable?
    \answerNA{}
  \item Did you describe any potential participant risks, with links to Institutional Review Board (IRB) approvals, if applicable?
    \answerNA{}
  \item Did you include the estimated hourly wage paid to participants and the total amount spent on participant compensation?
    \answerNA{}
\end{enumerate}

\end{enumerate}


\appendix

\section{Algorithm Detail}

About \textbf{Algorithm 2: Update Semantic Global Graph} compare Node Similarity. The similarity judgment method will greatly affect the number of $G_t^{SG}$ nodes, and then affect the semantic characteristics of Semantic Global Graph. We have added some changes to the node comparison method to make the final $G_t^{SG}$ more accurate. Please refer to Algorithm~\ref{Semantic_Global_Graph_Method_2} for details. The paragraphs marked in red are the newly added algorithm Jaccard Similarity. Jaccard Similarity can eliminate the need for repeated additions of nodes in adjacent frames. However, based on the results of the comparison experiment between Algorithm~\ref{Semantic_Global_Graph_Method_2} and Algorithm 2, we found that the effect is not significant.

\begin{algorithm}[H]
\SetAlgoLined
\DontPrintSemicolon
\SetKwInput{KwInput}{Input}
 \KwInput{$G_{t - 1}^{SG}, G_{t - 1}^{SC}, G_{t}^{SC}$}
  \tcp*[l]{Initialize}
  \If{t == 0} 
  {
      $Threshold$ set a constant value \tcp*[r]{$Threshold \in [0,1]$}
      $G_{0}^{SG} \in G^S = (V,E)$ \;
  }
  $G_{t}^{SG} \gets G_{t - 1}^{SG}$ \;
  \textcolor{red}{
    $jaccard\_similarity \gets \frac{G_{t}^{SC} \cap G_{t - 1}^{SC}}{G_{t}^{SC} \cup G_{t - 1}^{SC}} $ \tcp*[r]{Use Jaccard Similarity to calculate the similarity between $G_{t - 1}^{SC}$ and $G_{t}^{SC}$ }
  }
  \If{ \textcolor{red}{$jaccard\_similarity < 1$} } 
  {
      \textcolor{red}{
        $Candidate \: Object \gets Set(G_{t}^{SC}) - Set(G_{t - 1}^{SC})$ \;
      }
      \For{each $V, E$ in \textcolor{red}{$Candidate \: Object$} }
      {
        Use cosine similarity to calculate the similarity between $V$ 's node feature and $\forall V^{'}$ 's node feature \tcp*[r]{ $ V^{'} \in G_{t - 1}^{SG} $ }
        \If{similarity < $Threshold$}
        {
          $G_{t}^{SG}$ adds an new node and edge by $V, E$ \;
        }
      }
  }
  
 \KwRet $G_{t}^{SG}$ \;
 \caption{Update Semantic Global Graph (Method 2)}
 \label{Semantic_Global_Graph_Method_2}
\end{algorithm}

\section{Implementation Details}

Our experimental data is in Ubuntu 16.04.7 with two GeForce GTX 1080 Ti Intel(R) Xeon(R) CPU E5-2620 v4 @ 2.10GHz. Table~\ref{main_task_type} shows the detailed distribution of the data set. Due to the limitation of computing power, we choose to use the types of tasks to verify our proposed method.

\begin{table}[H]
  \caption{Main task type}
  \label{main_task_type}
  \centering
  \begin{tabular}{p{0.15\linewidth}  p{0.20\linewidth}  p{0.40\linewidth}}
    \toprule
    Main task type & Task distribution \newline train/val seen/val unseen & Low-dimensional basic operations \\
    \midrule
    Pick \& Place & 2842/112/113 & \multirow{7}{*}{\shortstack[l]{Stop, LookDown, LookUp,\\
                                                   MoveAhead, RotateLeft, RotateRight, \\
                                                   PickupObject, SliceObject, OpenObject, \\ 
                                                   PutObject, CloseObject, ToggleObjectOn, \\
                                                   ToggleObjectOff}} \\
    Stack \& Place & 2944/126/109 & \\
    Place Two & 2943/107/136 & \\
    Examine & 2251/94/173 & \\
    Heat \& Place & 3554/124/81 & \\
    Cool \& Place & 3244/115/109 & \\
    Clean \& Place & 3245/142/100 & \\
    \bottomrule
  \end{tabular}
\end{table}

\begin{figure}[H]
  \centering
  \includegraphics[width=0.9\linewidth]{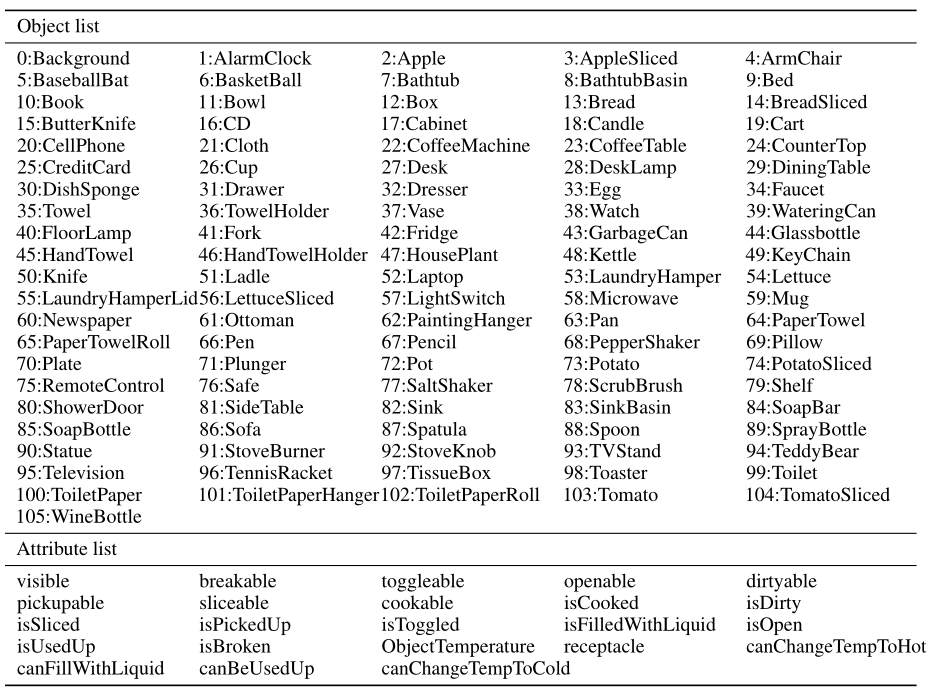}
  \caption{Object list \& Attribute list}
  \label{object_list_table}
\end{figure}

\textbf{Algorithm 3} details. The map created by Spatial Semantic Map $M_t \in \mathbb{R}^{s \text{x} s \text{x} c}$ is a square graph map with layers, $s=10$, $c=106$. The width of the map is 10, and the number of layers is 106. Layers and layers will be connected up and down. Only the nodes of the first layer will concatenate the neighbor nodes of the top, bottom, left, and right of the first layer, resulting in the relationship between the edges. Neighbor nodes between other layers are not connected. The objects represented by the number of layers are shown in Figure~\ref{object_list_table}.

There are a total of 106 different objects in the environment. As shown in Figure~\ref{object_list_table}, $feat^{V}$ and $feat^{word}$ of the object can be obtained through preprocessing. The features of $A Attribute \in \mathbb{R}^{1 \text{x} 23}$ are used to generate Semantic Prior Graph, which is an object attribute owned by AI2-Thor.

Table~\ref{ablation} shows the results of three Semantic Graph ablation. After adding Semantic Graph, most of the results can be improved.

\begin{table}[H]
  \caption{Semantic Graph ablation experiment. SG means Semantic Graph. Task means task success rate. }
  \label{ablation}
  \centering
  \begin{tabular}{p{0.1\linewidth} p{0.1\linewidth} p{0.1\linewidth} p{0.07\linewidth}  p{0.07\linewidth}  p{0.07\linewidth}  p{0.07\linewidth}  p{0.1\linewidth}}
    \toprule
    Priori SG & Current SG & Global SG & Task (\%) & Goto (\%) & Pickup (\%) & Put (\%) & Sub-Goal Avg. (\%) \\
    \midrule
    $\times$   &$\times$    &$\times$    & 2 & 39 & 64 & 7 & 36 \\
    $\times$   &$\times$    & \checkmark & 6 & 39 & 69 & 6 & 38 \\
    \checkmark &$\times$    &$\times$    & 6 & 38 & 67 & 7 & 37 \\
    \checkmark & \checkmark &$\times$    & 5 & 40 & 70 & 6 & 38 \\
    \checkmark & \checkmark & \checkmark & 8 & 48 & 67 & 5 & 40 \\
    \bottomrule
  \end{tabular}
\end{table}

\begin{figure}[H]
  \centering
  \includegraphics[width=0.9\linewidth]{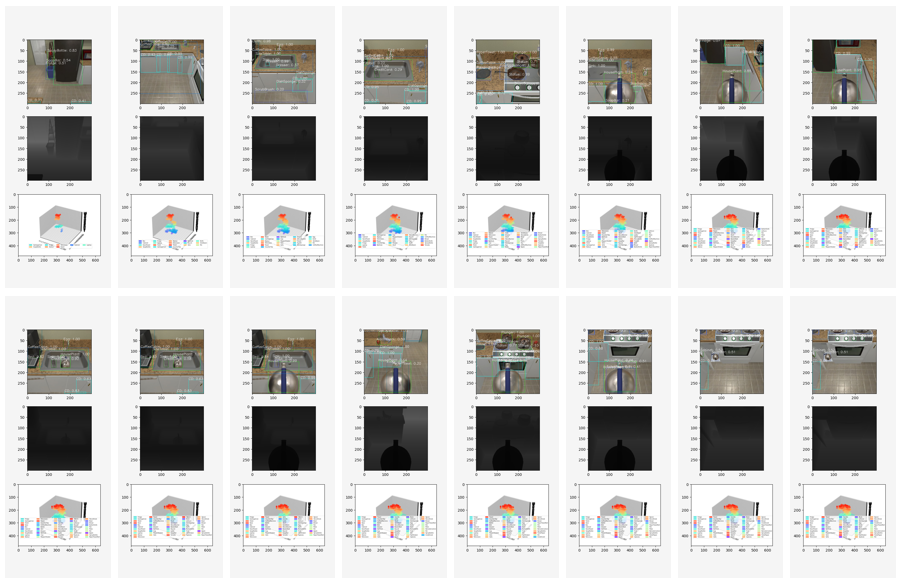}
  \caption{Display results of Spatial Semantic Map. The state of time changes from left to right.}
  \label{seq_graph_map}
\end{figure}

\begin{figure}[H]
  \centering
  \includegraphics[width=0.9\linewidth]{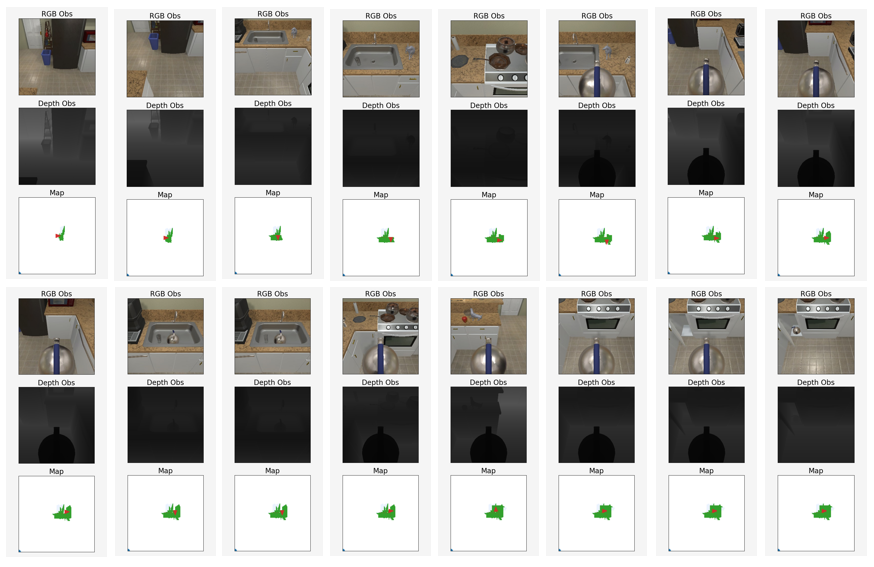}
  \caption{Display results of SLAM Map. The state of time changes from left to right.}
  \label{seq_slam_map}
\end{figure}

Figure~\ref{seq_graph_map} and Figure~\ref{seq_slam_map} show the detailed changes recorded by Spatial Semantic Map and SLAM during task execution.

\end{document}